\pdfoutput=1
\documentclass[11pt]{article}

\usepackage[final]{coling}

\usepackage{times}
\usepackage{latexsym}
\usepackage{amsmath}

\usepackage{algorithm} 
\usepackage[algo2e,ruled]{algorithm2e} 
\usepackage[]{algpseudocode}

\usepackage{subcaption}
\usepackage{svg}
\usepackage{booktabs}
\usepackage{multirow}

\newcommand{\Ccal}{\mathcal{C}}
\newcommand{\Dcal}{\mathcal{D}}
\newcommand{\Ecal}{\mathcal{E}}

\newcommand{\Mcal}{\mathcal{M}}
\newcommand{\Ncal}{\mathcal{N}}

\newcommand{\Pcal}{\mathcal{P}}
\newcommand{\Qcal}{\mathcal{Q}}

\newcommand{\Ucal}{\mathcal{U}}

\newcommand{\Xcal}{\mathcal{X}}

\newcommand{\kmeans}{$k$-\textsc{MEANS}}
\newcommand{\kmeansplusplus}{$k$-\textsc{MEANS++}}

\urldef{\cfourdataset}\url{https://www.tensorflow.org/datasets/catalog/c4}

\urldef{\bartmodelhuggingface}\url{https://huggingface.co/facebook/bart-base}

\urldef{\altoolbox}\url{https://github.com/AIRI-Institute/al_toolbox}

\urldef{\unbabelcomet}\url{https://github.com/Unbabel/COMET}

\urldef{\huggingfacedatasets}\url{https://huggingface.co/datasets?task_categories=task_categories:translation}
\urldef{\huggingfacemodels}\url{https://huggingface.co/models?pipeline_tag=translation}

\usepackage[T1]{fontenc}
\usepackage[utf8]{inputenc}

\usepackage{microtype}

\usepackage{inconsolata}

\title{To Label or Not to Label: Hybrid Active Learning \\ for Neural Machine Translation}

\author{Abdul Hameed Azeemi \and Ihsan Ayyub Qazi \and Agha Ali Raza \\
         Lahore University of Management Sciences \\ Pakistan \\
 \texttt{\{abdul.azeemi, ihsan.qazi, agha.ali.raza\}@lums.edu.pk} }
\begin{document}
\maketitle

\begin{abstract}
Active learning (AL) techniques reduce labeling costs for training neural machine translation (NMT) models by selecting smaller representative subsets from unlabeled data for annotation. Diversity sampling techniques select heterogeneous instances, while uncertainty sampling methods select instances with the highest model uncertainty. Both approaches have limitations - diversity methods may extract varied but trivial examples, while uncertainty sampling can yield repetitive, uninformative instances. To bridge this gap, we propose Hybrid Uncertainty and Diversity Sampling (HUDS), an AL strategy for domain adaptation in NMT that combines uncertainty and diversity for sentence selection. HUDS computes uncertainty scores for unlabeled sentences and subsequently stratifies them. It then clusters sentence embeddings within each stratum and computes diversity scores by distance to the centroid. A weighted hybrid score that combines uncertainty and diversity is then used to select the top instances for annotation in each AL iteration. Experiments on multi-domain German-English and French-English datasets demonstrate the better performance of HUDS over other strong AL baselines. We analyze the sentence selection with HUDS and show that it prioritizes diverse instances having high model uncertainty for annotation in early AL iterations.
\end{abstract}

\section{Introduction}

\begin{figure}[t]
\begin{center}
\centerline{\includegraphics[width=\columnwidth]{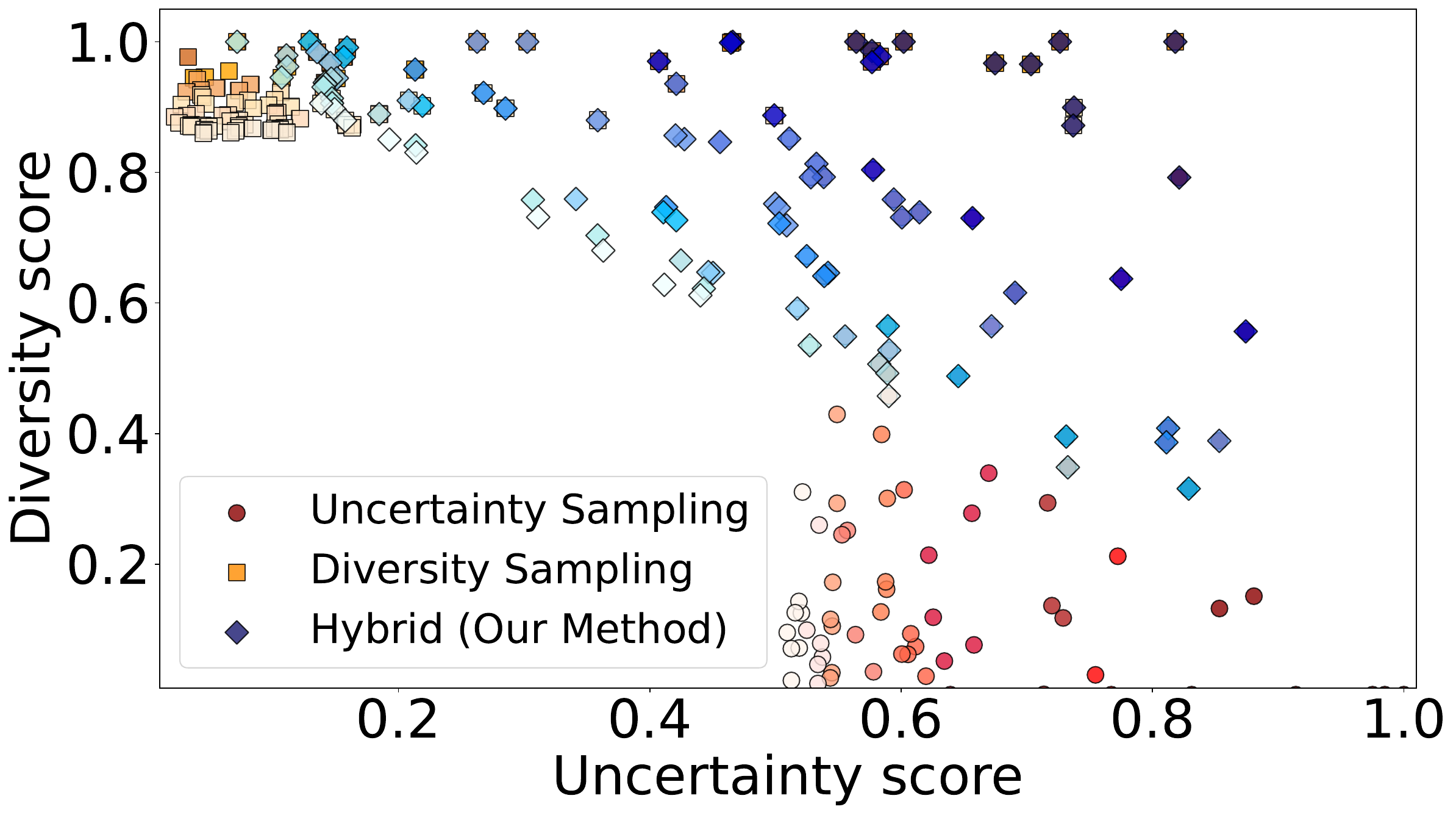}}
\caption{Comparison of the instances selected through different active learning strategies on the task of English to German (En $\rightarrow$ De) translation using the Law domain dataset and \texttt{BART-base} model. The color represents the active learning iteration; from the darkest (first iteration) to the lightest (tenth iteration). Our proposed hybrid active learning method (\textsc{HUDS}) selects diverse yet challenging sentences for annotation in early iterations. }
\label{fig:mainimage}
\end{center}
\vskip -0.3in
\end{figure}

\noindent The training process for Machine Translation (MT) models typically requires large, high-quality parallel corpora to achieve robust performance. However, acquiring such data is costly and laborious, especially for low-resource languages and domain-specific data which requires professional translators with domain knowledge. For instance, training an MT model to translate specific medical conditions requires extensive labeling by medicine experts. Active learning (AL) is a well-known strategy that helps to reduce the labeling requirements by selecting a smaller representative subset for annotation, thereby reducing the overall cost.  AL follows an iterative procedure that involves (1) querying instances from an unlabeled pool; (2) obtaining annotations for unlabeled instances through a human expert; and (3) adding the annotated examples to the labeled dataset and retraining. 

In the context of Neural Machine Translation (NMT), active learning aids in selecting a small number of sentences for labeling that are likely to bring a similar performance if the NMT model is trained on a larger labeled set. AL methods for NMT can be broadly classified into diversity and uncertainty sampling methods, often described as the ``two faces of active learning'' \cite{dasgupta2011two}. Diversity sampling ensures that a heterogeneous set of instances from the unlabeled set is selected for annotation. In contrast, uncertainty sampling attempts to select the instances with the highest model uncertainty, i.e., the ones which are the most difficult for the next AL iteration. Both these methods have their limitations, such that the methods focusing on diversity can extract varied but trivial examples, whereas uncertainty sampling may lead to the selection of instances with high uncertainty but which are repetitive and not useful. To mitigate this, hybrid sampling approaches have been proposed for other NLP tasks, including named entity recognition \cite{kim-etal-2006-mmr}, abstractive text summarization \cite{tsvigun2023active}, and text classification \cite{yuan-etal-2020-cold}. No hybrid AL strategy for efficiently acquiring domain-specific data in NMT has been proposed yet which successfully incorporates model uncertainty and data diversity in the sampling procedure.

\begin{figure*}[!ht]
\begin{center}
\centerline{\includegraphics[width=0.95\textwidth]{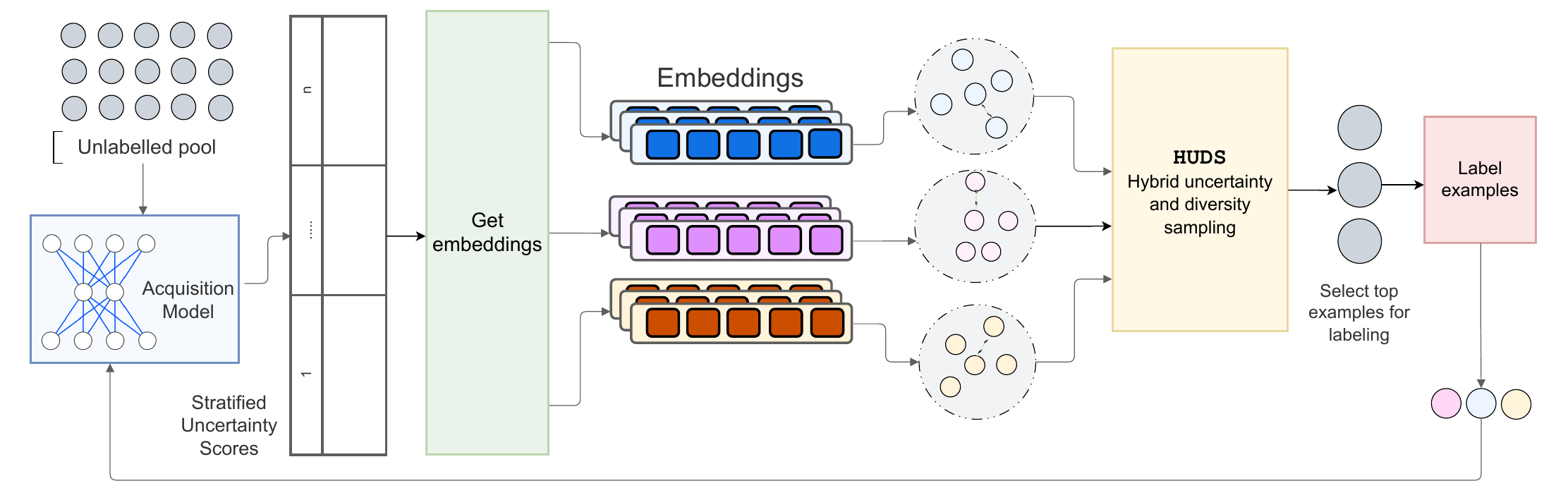}}
\caption{A representation of hybrid uncertainty and diversity sampling (HUDS) for active learning in NMT.  }
\label{fig:huds}
\end{center}
\vskip -0.2in
\end{figure*}

To develop a hybrid AL strategy for NMT, we should consider a multi-step procedure involving uncertainty computation, diversity sampling, and a weighted hybrid acquisition function that leverages both. In this work, we develop \textsc{HUDS} (Hybrid Uncertainty and Diversity Sampling), a hybrid active learning strategy for NMT. We first compute uncertainty scores for unlabeled sentences and stratify them. We then obtain embeddings for each sentence in a stratum, cluster them using \kmeans{} and compute the diversity score for each instance by their distance to the centroid. Finally, we compute a hybrid sampling score through a weighted sum of the uncertainty and diversity scores and select the \textit{top}-k instances with the highest scores for annotation in each iteration of AL. This weighted sampling ensures the selection of diverse yet challenging instances for annotation (Fig. \ref{fig:mainimage}), which contributes to the model's robust performance on domain-specific datasets.

\subsection{Contributions}
\begin{itemize}
    \item To the best of our knowledge, we present the first hybrid active learning strategy (\textsc{HUDS}) for NMT.
    \item We evaluate \textsc{HUDS} on German-English and French-English datasets spanning different domains (Medicine, Law, IT, Ted talks) with the \texttt{BART} language model and demonstrate that \textsc{HUDS} outperforms other AL baselines in NMT.
    \item We examine the examples selected for annotation by \textsc{HUDS} and demonstrate that it prioritizes diverse yet challenging instances for annotation in early AL iterations. We also find that \textsc{HUDS} selects varied instances with higher unigram coverage compared to other AL strategies.
\end{itemize}

\section{Preliminaries}

\subsection{Neural Machine Translation}
The primary objective of a Neural Machine Translation (NMT) model is to convert a source sentence $\mathbf{X}=\left\{x_{1}, \ldots, x_{S}\right\}$ into a corresponding target sentence $\mathbf{Y}=\left\{y_{1}, \ldots, y_{T}\right\}$, where $\mathbf{X}$ and $\mathbf{Y}$ each contain $S$ and $T$ tokens, respectively. The likelihood of each token in the target sentence, given the source sentence, is depicted using the chain rule below:
\begin{equation}
P(\mathbf{Y} \mid \mathbf{X} ; \theta)=\prod_{i=1}^{T} p\left(y_{i} \mid y_{0:i-1}, \mathbf{X} ; \theta\right)
\end{equation}
In the above equation, $\theta$ symbolizes the parameters of the model. NMT models strive to optimize the cross-entropy (CE) loss, achieving this by decreasing the negative log-likelihood of the training examples, represented as:
\begin{equation}
\mathcal{L}_{\mathrm{CE}}(\theta)=-\sum_{i=1}^{T} \log p\left(y_{i} \mid y_{0:i}, \mathbf{X} ; \theta\right)
\label{eq:celoss}
\end{equation}

When the model performs inference, it produces probabilities for the target tokens via an auto-regressive process. These probabilities then aid in choosing high-probability tokens with the help of search strategies such as beam search.

\subsection{Active Learning for NMT} We consider a labeled corpus $\mathcal{D}_{l}$ and an unlabeled corpus $\mathcal{D}_{u}$ for active learning in NMT. In the context of domain adaptation, $\mathcal{D}_{l}$ is a larger out-of-domain labeled corpus (e.g., WMT14 German-English corpus), and $\Dcal_{u}$ is a small in-domain unlabeled corpus (e.g., Medicine domain German-English corpus). For acquiring the translation of the sentences selected by the query strategy, we consider a human oracle who can translate any sentence in the unlabeled data. We consider a fixed budget $c$ for annotation such that all the annotation through must occur within that budget.

\subsection{Contrasting Uncertainty and Diversity}
Active learning strategies can be broadly classified into diversity-sampling methods and uncertainty-sampling methods, also known as "two faces of active learning" \cite{dasgupta2011two}. Diversity sampling methods aim to select diverse instances for annotation, whereas uncertainty sampling techniques use the model's uncertainty as a proxy to select the hardest examples that should be annotated. To achieve optimal results, any robust AL strategy should leverage both uncertainty and diversity. However, designing such a hybrid approach is not trivial \cite{hsu2015active} and is an open research problem \cite{yuan-etal-2020-cold}.

We now discuss two promising hybrid active learning strategies \textsc{Badge} \cite{ash2019deep} and \textsc{ALPS} \cite{yuan-etal-2020-cold} for classification tasks that combine the benefits of uncertainty and diversity sampling. Inspired by these two strategies, we then present a new hybrid active learning strategy for the task of neural machine translation.

\subsubsection{\textsc{Badge}}
The primary objective of the \textsc{BADGE} algorithm is to select heterogeneous and uncertain instances for annotation in each iteration of AL. This is achieved by clustering special representations called \textit{gradient embeddings} that encapsulate model certainty. Once the gradient embeddings are computed, \textsc{BADGE} uses \kmeansplusplus{} seeding algorithm \cite{arthur2007k} to select a subset of examples $S_t$ for annotation. The examples included in this subset are diverse in terms of uncertainty and data distribution due to the way gradient embeddings are calculated.

\subsubsection{\textsc{ALPS}}
Improving upon \textsc{Badge}, \textsc{Alps} attempts to estimate the uncertainty of instances using the masked language model (MLM) loss of a pre-trained BERT within the downstream classification task. Thus, \textsc{Alps} works in a cold-start setting by leveraging the loss from a pre-trained BERT in contrast to \textsc{Badge} which estimates uncertainty using the classification model that is being trained, i.e, in a warm-start setting. \textsc{Alps} achieves this by creating a surprisal embedding $s_x$ for each input $x$. This process involves feeding the non-masked input $x$ into the BERT MLM head and calculating the cross-entropy loss for 15\% randomly selected tokens. These embeddings are then clustered and a diverse subset of examples are selected for the next iteration.

\begin{algorithm*}[!ht]
	\caption{Single iteration of \textsc{HUDS} \label{algo:huds}}

    \DontPrintSemicolon
    \SetAlgoLined

	\KwIn{Unlabeled pool $\Dcal_{u}$, Number of strata $n$, Pre-trained encoder $\mathcal{P}$, Acquisition Model $\Mcal$, Hybrid parameter $\lambda$, Instances to select $k$ }
    $\Ucal_{s} \leftarrow \mathcal{M}(u)$, $u \in \Dcal_{u}$ \Comment{Compute uncertainty scores for each sentence in $\Dcal_{u}$}\;

    $s_{min} \leftarrow min(\Ucal_{s})$, $s_{max} \leftarrow max(\Ucal_{s})$, $r \leftarrow s_{max} - s_{min}$, $\Ncal \leftarrow \emptyset $\;

    \For{$i = 1 \to n$} {
    $\Ncal \leftarrow  \Ncal \cup \Ucal_{s}\left(s_{\text{min}} + \frac{i-1}{n}(r), s_{\text{min}} + \frac{i}{n}(r)\right)$  \Comment{Perform stratification on uncertainty scores $\Ucal_{s}$}\;
    }

    $\Ccal \leftarrow \emptyset$\;

	\For {$\Xcal$ in $\Ncal$}{

    $\Ecal \leftarrow \emptyset$\;
    
    \For {sentences $x \in \Xcal$} {
        $\Ecal \leftarrow \Ecal \cup \Pcal(x)$ \Comment{Obtain embeddings for each sentence in stratum $\Xcal$}
    }
    $c \leftarrow \kmeans{}(\Ecal)$\; \Comment{Perform \kmeans{} clustering on embeddings}

    $\Ccal \leftarrow  \Ccal \cup \lambda \cdot d(c, e) + (1 - \lambda) \cdot \Ucal_{s}(e) , e \in \Ecal $\; \Comment{Compute hybrid score for each sentence}
	}
    $\Qcal \leftarrow topK(\Ccal, k)$\Comment{Select top k instances for annotation}

	\KwOut{ $\Qcal$ }
\end{algorithm*}

\section{Hybrid Uncertainty and Diversity Sampling}
We now present a novel hybrid strategy for active learning in NMT named hybrid uncertainty and diversity sampling (\textsc{HUDS}). HUDS is a multi-step procedure involving uncertainty computation, diversity sampling, and a weighted hybrid acquisition function that leverages both of these. The complete workflow is shown in Fig. \ref{fig:huds}.

\medskip
\noindent \textbf{Uncertainty scores.} We first estimate the uncertainty for an unlabeled sentence $s$ through the normalized negative log-likelihood: $NNLL(X) = -\frac{1}{S} \sum_{j=1}^{S} \log p(x_j | x_1, x_2, \dots, x_{j-1})$, where $S$ is the sentence length and $x_j$ is the $j^{th}$ token in $X$. Lower NLL values indicate higher model confidence in the prediction, while higher values indicate less confidence thus greater uncertainty. We then perform stratification on uncertainty scores, with the range for $i$th stratum defined as,
\begin{equation}
s_{i} = \left(s_{min}+\frac{i-1}{n}\left(r\right), s_{min}+\frac{i}{n}\left(r\right)\right)
\end{equation}
where $s_{min}$ is the minimum uncertainty, $s_{max}$ is the maximum uncertainty, $r$ is the range specified by $s_{max}-s_{min}$, and $n$ is the total number of strata.

\medskip
\noindent \textbf{Diversity scores.} We then utilize a pre-trained BERT to obtain the embeddings of sentences within each stratum. Subsequently, we cluster the embeddings using \kmeans{} and compute the diversity score for each instance by the cosine distance between that instance and the cluster centroid. The goal of clustering is only to determine the distance of each example from the centroid of the (already defined) stratum and thus a value of $k = 1$ suffices here. In other words, diversity is represented with distance from the average vector, over sentences grouped by uncertainty strata. These diversity scores represent the heterogeneity of the examples. Therefore, the semantically dissimilar (and potentially harder) instances in each stratum have higher scores than the prototypical examples. 

\medskip
\noindent \textbf{Hybrid scores.} We compute a hybrid sampling score $H(x)$ for a sentence $x$ through the weighted sum of the uncertainty and diversity scores (Eq. \ref{eq:huds}) and select the \textit{top}-k instances with the highest scores for annotation.
\begin{equation}
H(x) = \lambda \cdot d(x, c_{i}) + (1 - \lambda) \cdot u_{x}
\label{eq:huds}
\end{equation}
where $d(x, c_{i})$ is the diversity score represented by cosine distance between the embedding of sentence $x$ and the centroid of the cluster $c_{i}$ of stratum $s_{i}$ (to which the sentence $x$ belongs), while $u_{x}$ is the uncertainty score of sentence $x$. Algorithm \ref{algo:huds} summarizes the complete procedure.

Stratification of uncertainty scores allows the selection of diverse sentences from each subpopulation, i.e., examples with a low uncertainty can also be selected as long as they are diverse. This is in contrast to the other pure uncertainty sampling methods which prevent the selection of informative sentences that have a low uncertainty. We hypothesize that hybrid sampling aids in selecting sentences with low overall uncertainty but having harder (dissimilar) segments that the NMT model cannot translate well.

\section{Experimental Setup}
We mirror the conventional active learning setup followed in previous studies \cite{shelmanov_active,dor2020active, hu-neubig-2021-phrase, tsvigun2023active}. We first train the NMT model on the out-of-domain labeled corpus. We then start the AL procedure with a randomly selected small subset of labeled sentences from in-domain data and in the first iteration fine-tune the NMT model on this subset. In each subsequent iteration, we use the query strategy $\Qcal$ to select the examples for annotation, add them to the annotated 
pool of examples (i.e., emulate the manual labeling process of translation without human involvement), and remove them from the unlabeled set. The NMT model is then trained on the sentence pairs in the labeled pool and evaluated on an unseen validation set. We use \textsc{SacreBLEU} \cite{post-2018-call} to evaluate the performance on the validation set in each iteration, according to the protocol followed in previous works on AL in NMT \cite{hu2021phrase, zhao2020active}. \textsc{SacreBLEU} uses the standard WMT tokenization, addresses several issues in the reporting of regular \textsc{BLEU} scores, and is more reproducible \cite{post-2018-call}.

\subsection{Datasets}

We utilize the WMT14 En-De (English-German) parallel data as our out-of-domain labeled dataset. It contains 4.5M training pairs and 3000 validation and test pairs. For the in-domain data, we use the multi-domains dataset containing German-English parallel data for five domains \cite{koehn-knowles-2017-six}, deduplicated and resplit by \citet{aharoni-goldberg-2020-unsupervised}. This dataset has been frequently used for the evaluation of domain adaptation and active learning approaches in neural machine translation \cite{dou19emnlp, hu-neubig-2021-phrase}. We use three corpora for our experiments: Medicine, Law, and IT. The statistics are shown in Table \ref{tab:datastatistics}.

\begin{table}[!t]
\centering
\resizebox{\columnwidth}{!}{
\begin{tabular}{lllll}
\hline Dataset & \# Train Pairs & \# Val Pairs & \# Test Pairs \\
\hline Medicine & 248,099 & 2,000 & 2,000 \\
Law & 467,309 & 2,000 & 2,000 \\
IT & 222,927 & 2,000 & 2,000 \\
\hline
\end{tabular}}
\caption{Data statistics of the multi-domains German-English corpora in
the Medicine, IT, and Law domains.} 
\label{tab:datastatistics}
\end{table}

\subsection{Model}
We use the \texttt{BART} model\footnote{\bartmodelhuggingface} for our experiments which works well for natural language generation tasks including translation \cite{lewis-etal-2020-bart}. \texttt{BART} is a transformer encoder-decoder model having a bidirectional encoder and an autoregressive decoder. Our choice of  \texttt{BART} is motivated by recent works on active learning for text generation that have employed encoder-decoder architectures such as BART and demonstrated their effectiveness \cite{xia-etal-2024-hallucination, tsvigun2023active}. We did not use a large language model (LLM) for the domain adaptation experiments, since LLMs are typically pre-trained on massive, diverse corpora that often overlap with the test domain. Such pre-training raises the risk of data contamination \cite{sainz-etal-2023-nlp}, making it difficult to accurately assess the model’s ability to learn from newly introduced, domain-specific data. As a result, improvements observed in an LLM would not reliably reflect genuine gains from the active learning process, but rather the model’s prior exposure to similar samples.

We conduct experiments using the \texttt{BART-base} variant pre-trained on Wikipedia and the BookCorpus. This variant has 140 million parameters and six blocks in the encoder and decoder. We train the model using the AdamW optimizer with a training batch size of 16. The learning rate is set to 2e-5 with a weight decay of 0.028 and gradient clipping of 0.28. The beam size is set to 4 for evaluation. We use $\lambda = 0.5$ for HUDS sampling (after the first query iteration), giving equal weightage to uncertainty and diversity in the computation of hybrid scores. Uncertainty is computed for 20,000 randomly selected examples from the unlabeled pool as it is computationally expensive to be performed for the complete dataset. 1000 sentences are queried in each iteration. The experiments are run on a single 48GB NVIDIA A6000 GPU.

\begin{table*}[t]
\begin{center}
\begin{small}

\resizebox{0.9\textwidth}{!}{%
\begin{tabular}{lrrrrrrr}
\toprule
\textsc{\shortstack{Dataset}} &  \textsc{AL}  & \multicolumn{5}{c}{\textsc{AL Iteration}} \\ 
  & \textsc{Strategy} & 2 & 4 & 6 & 8 & 10  \\ 

\midrule

\multirow{5}{*}{\shortstack{\textsc{Medicine}}}  & Random & 29.36$_{\pm0.1}$  & 30.75$_{\pm0.1}$  & 31.44$_{\pm0.1}$  & 31.93$_{\pm0.1}$ &  32.70$_{\pm0.1}$   \\
 &  NSP & 29.69$_{\pm0.6}$  &  31.39$_{\pm0.5}$  & 32.27$_{\pm0.4}$  & 32.89$_{\pm0.3}$  & 33.68$_{\pm0.2}$ \\
 & IDDS & 29.45$_{\pm0.1}$  & 31.11$_{\pm0.1}$  & 32.08$_{\pm0.3}$  & 32.76$_{\pm0.2}$  & 33.05$_{\pm0.0}$ \\
 & \textsc{HUDS} & \textbf{30.76}$_{\pm0.2}$  & \textbf{32.03}$_{\pm0.1}$  &  \textbf{33.14}$_{\pm0.1}$  & \textbf{33.65}$_{\pm0.1}$  & \textbf{34.58}$_{\pm0.1}$ \\

\midrule

\multirow{5}{*}{\shortstack{\textsc{IT}}}  & Random & 27.64$_{\pm0.3}$  & 28.30$_{\pm0.1}$  & 28.78$_{\pm0.1}$  & 29.28$_{\pm0.1}$ &  29.54$_{\pm0.1}$   \\
 &  NSP & 28.43$_{\pm0.2}$  &  29.26$_{\pm0.1}$  & 29.70$_{\pm0.2}$  & 30.09$_{\pm0.1}$  & 30.23$_{\pm0.2}$ \\
 & IDDS & 23.81$_{\pm0.1}$  & 24.31$_{\pm0.2}$  & 24.84$_{\pm0.0}$  & 25.21$_{\pm0.1}$  & 25.41$_{\pm0.1}$ \\
 & \textsc{HUDS} & \textbf{29.15}$_{\pm0.1}$  & \textbf{29.69}$_{\pm0.1}$  &  \textbf{30.12}$_{\pm0.1}$  & \textbf{30.54}$_{\pm0.1}$  & \textbf{30.70}$_{\pm0.0}$ \\

\midrule

\multirow{5}{*}{\shortstack{\textsc{Law}}}  & Random & 32.60$_{\pm0.3}$  & 33.52$_{\pm0.5}$  & 34.28$_{\pm0.4}$  & 34.80$_{\pm0.5}$ &  35.21$_{\pm0.5}$   \\
 &  NSP & 32.89$_{\pm0.1}$  &  34.18$_{\pm0.0}$  & 34.73$_{\pm0.1}$  & 35.16$_{\pm0.3}$  & 35.53$_{\pm0.0}$ \\
 & IDDS & 32.91$_{\pm0.0}$  & 34.12$_{\pm0.1}$  & 34.61$_{\pm0.1}$  & 35.07$_{\pm0.0}$  & 35.33$_{\pm0.0}$ \\
 & \textsc{HUDS} & \textbf{33.76}$_{\pm0.0}$  & \textbf{34.39}$_{\pm0.2}$  &  \textbf{34.89}$_{\pm0.1}$  & \textbf{35.17}$_{\pm0.0}$  & \textbf{35.62}$_{\pm0.1}$ \\

\bottomrule
\end{tabular}
}
\end{small}
\end{center}

  \caption{\textsc{SacreBLEU} scores with different active learning strategies for NMT across different domain datasets. \texttt{BART-base} model is used in all experiments. Active learning is done over ten iterations with 1000 sentences queried in each iteration. For each strategy, three independent runs are done with different seeds, and the mean \textsc{SacreBLEU} is reported, with the subscript representing the standard error. Best results are given in \textbf{bold. }HUDS consistently shows better performance compared to other AL strategies.}
  \label{tab:mainresults}
\end{table*}

\subsection{Baselines}
AL strategies for NMT primarily include a sentence-level selection of examples or phrase-level selection for training the NMT model. Sentence-level strategies focus on selecting \textit{sentences} that are most useful for training while phrase-level strategies choose individual phrases that are the most informative. In some scenarios, phrase-level strategies can be more cost-effective as they provide granular control to avoid the inclusion of otherwise informative sentences which have phrases that the NMT model can already translate well. However, their implementation for NMT is generally more complex than sentence-level selection, involving the creation of synthetic parallel data that incorporates phrasal translations followed by a data mixing procedure that is utilized for mixed fine-tuning \cite{hu-neubig-2021-phrase}. Thus, we compare our approach to three sentence-level selection baselines, including Random selection, Normalized Sequence Probability (NSP) \cite{ueffing_sp}, and In-Domain Diversity Sampling (IDDS) \cite{tsvigun2023active}. NSP and IDDS are strong baselines, while random selection is competitive in AL given a sufficiently large annotation budget \cite{hu2021phrase}. 

While BADGE and ALPS are promising AL methods for classification tasks, applying them to NMT requires significant modification to the acquisition functions. Thus, we do not consider them as baselines for AL in NMT. Additionally, we do not include subset selection approaches as baselines, e.g., \citet{kothawade2022prism}, as their goal is to select a subset from a \textit{fixed} set of (often \textit{fully labeled}) examples to allow training in resource-constrained environments. In contrast, active learning aims to iteratively select a subset of examples for labeling from an \textit{unlabeled} pool, with the aim of minimizing the labeling cost.

\medskip
\noindent \textbf{Random Sampling.} Random sampling involves selecting a subset of sentences randomly from the unannotated data for labeling. Despite its simplicity, it has shown strong performance in active learning for various NLP tasks \cite{miura2016selecting, tsvigun2023active} as it reflects the distribution of the complete unlabeled set without bias. Moreover, it is an efficient strategy as no queries are needed for example selection in contrast to the other AL strategies.

\medskip
\noindent \textbf{Normalized Sequence Probability (NSP).} This is an uncertainty-based query strategy that helps to quantify the uncertainty of a model's predictions on a particular sequence. It was introduced by \citet{ueffing_sp} and is a strong baseline that is the basis for several other AL techniques \cite{haffari2009active, tsvigun-etal-2022-altoolbox}. NSP for a sentence $\mathbf{v}$ is defined as $1-\bar{p}_{\hat{\mathbf{w}}}(\mathbf{y} \mid \mathbf{v})$, where $\bar{p}_{\hat{\mathbf{w}}}(\mathbf{y} \mid \mathbf{v})$ is the geometric mean of the token probabilities predicted by model.

\medskip
\noindent \textbf{In-Domain Diversity Sampling (IDDS).} This strategy \cite{tsvigun2023active} attempts to select sentences that are different from the previously annotated ones in an attempt to increase the diversity of the final set. Simultaneously, it prevents the selection of noisier instances (or outliers) that are significantly different semantically from the documents in the in-domain dataset. This mitigates the prevalent issue of selecting noisier instances found in other uncertainty-based AL strategies. The acquisition function of IDDS is given by:

\vskip -0.1in
\begingroup
\small
\begin{equation}
IDDS(\mathbf{v}) = \alpha \frac{\sum\limits_{k = 1}^{|A|} f(\mathbf{v}, \mathbf{v}_k)}{|A|} 
- (1 - \alpha) \frac{\sum\limits_{m = 1}^{|B|} f(\mathbf{v}, \mathbf{v}_m)}{|B|}
\label{eq:idds}
\end{equation}
\endgroup

In this equation, $f(\mathbf{v}, \mathbf{v}')$ is a function measuring text similarity, $B$ represents the labeled set,  $A$ represents the unlabeled set, and $\alpha$ is a hyperparameter between 0 and 1. IDDS demonstrates strong results on the Abstractive Text Summarization task, outperforming other AL strategies. Thus, we select this strategy as one of the baselines for comparison.

\begin{figure*}[!ht]
\begin{center}
\centerline{\includegraphics[width=\textwidth]{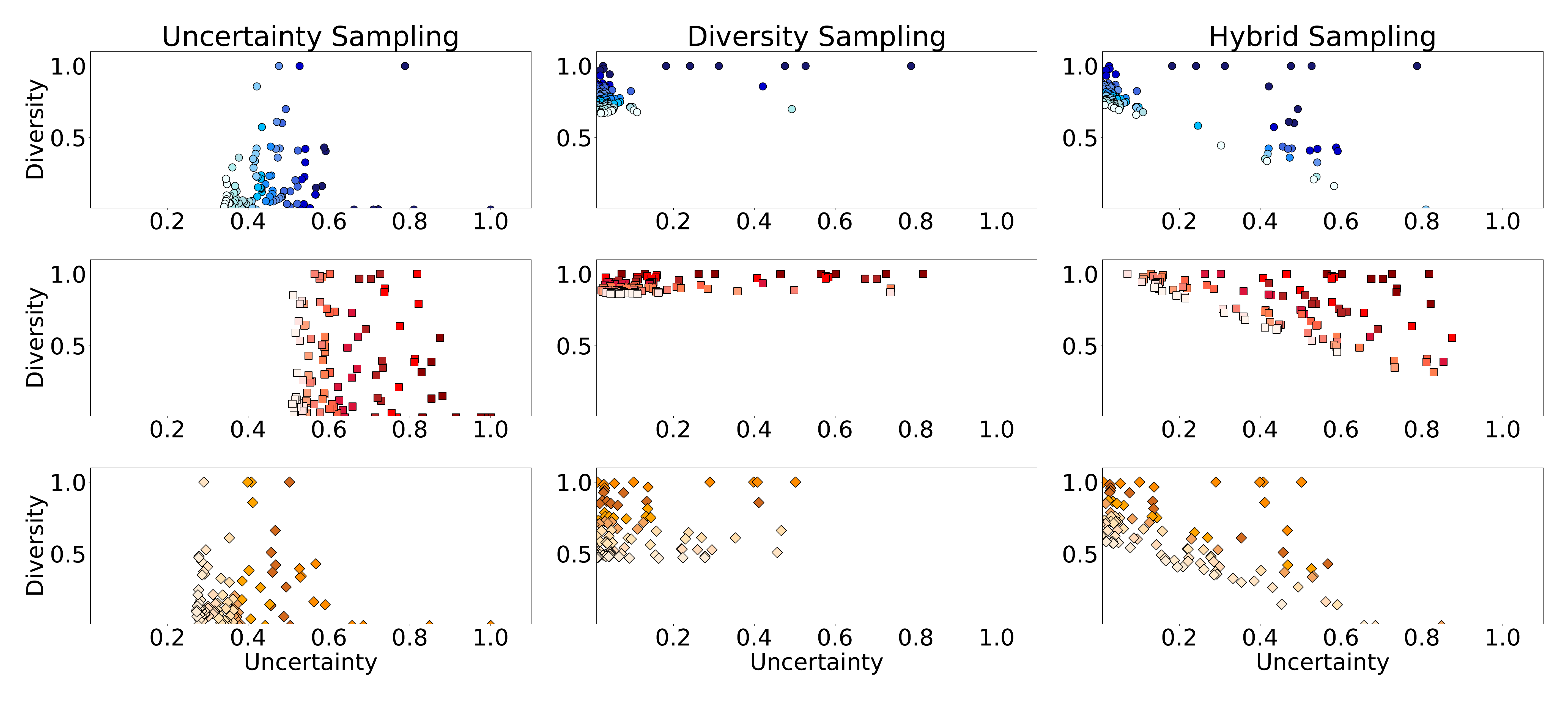}}
\caption{Plot of uncertainty vs diversity scores for Medicine (\textit{Row 1}), Law (\textit{Row 2}), and IT (\textit{Row 3}) datasets. Each plot contains 100 markers representing sentences that were selected during the AL procedure over 10 iterations, with 10 sentences selected in each iteration. The color represents sampling iteration in AL, with the darkest color representing the first iteration and the lightest color representing the tenth iteration. Columns correspond to the sampling strategy. Across all three datasets, hybrid sampling tends to select sentences with high uncertainty and diversity.  }
 \label{fig:samplingcomparison}
\end{center}
\vskip -0.4in
\end{figure*}

\section{Results}

Table \ref{tab:mainresults} shows the BLEU scores of \texttt{BART-base} model fine-tuned with different AL strategies across three datasets. We do three runs with different random seeds for each selection strategy and dataset and report the mean \textsc{SacreBLEU}. We observe that \textsc{HUDS} performs better than random sampling, NSP, and IDDS across all iterations. For the medicine dataset, HUDS demonstrates 34.58 \textsc{SacreBLEU} at the end of the final iteration compared to 33.68 for NSP, 33.05 for IDDS, and 32.70 for Random. The differences in the  \textsc{SacreBLEU} scores are statistically significant by Wilcoxon’s signed-rank test (p < 0.05). We also notice that the relative ranking of \textsc{SacreBLEU} for different strategies varies significantly across datasets. We hypothesize that this indicates a difference in the distribution and, more specifically, the uncertainty and diversity of instances that we uncover next. Interestingly, IDDS does not perform well on the IT dataset, with a 13.8\% lower \textsc{SacreBLEU} compared to random sampling (25.4 vs 29.5). We conjecture that this is due to the selection of diverse instances in each iteration that are different from the already annotated ones but are uninformative for the model.

\subsection{Analyzing \textsc{HUDS}}
\label{section:analyzinghuds}
To understand the sentence selection preference in HUDS during AL iterations, it is important to find out how it compares to regular uncertainty and diversity sampling. We plot uncertainty vs. diversity scores (Fig. \ref{fig:samplingcomparison}) and analyze the unlabeled instances selected in ten iterations for annotation by each AL strategy (uncertainty, diversity, \textsc{HUDS}) across three datasets. In the earlier iterations (represented by dark-colored shapes), \textsc{HUDS} selects instances with high values of uncertainty and diversity for Medicine and IT datasets. For the Medicine and Law datasets, a few diverse instances have moderate to low uncertainty (Row 1 and 3, Column 2 of Fig. \ref{fig:samplingcomparison}); hence \textsc{HUDS} selects those as well. In contrast, uncertainty sampling always selects uncertain instances regardless of their diversity, and diversity sampling prefers heterogeneous instances only. The difference in the distribution of uncertainty and diversity across datasets also helps explain the difference in the relative ranking of strategies in terms of \textsc{SacreBLEU} in Table \ref{tab:mainresults}.

\subsection{Evaluation on En-Fr corpus}

\begin{table}[H]
\centering

\resizebox{0.8\columnwidth}{!}{
\begin{tabular}{c c c c}
%\small
\hline
\textbf{Random} & \textbf{IDDS} & \textbf{NSP} & \textbf{HUDS} \\
\hline
34.8  & 33.5 & 35.1 & \textbf{35.5} \\
\hline
\end{tabular}}
\caption{Comparison of the mean \textsc{SacreBLEU} after the final AL iteration on the IWSLT 2014 En-Fr validation set.}
\label{tab:enfresacrebleu}
\vskip -0.2in
\end{table}

To confirm the generalization of HUDS to other languages, we conduct an experiment on the English-French (En-Fr) language pair. We utilize the WMT14 (En-Fr) parallel data as the out-of-domain labeled dataset and the IWSLT 2014 Ted Talks (En-Fr) as the in-domain dataset, following the protocol presented in \citet{wang2017sentence}. We use the \texttt{BART-base} model for AL. HUDS demonstrates the highest \textsc{SacreBLEU} score on the IWSLT 2014 validation set compared to other strategies, confirming its generalization to En-Fr (Table \ref{tab:enfresacrebleu}).

\subsection{Changing annotation size}
To determine the impact of a larger annotation size on the translation quality, we conduct an experiment on the IT dataset with 5000 examples labeled in each iteration (instead of 1000 in the earlier experiments). Ten iterations are performed, which brings the final quantity of selected sentences to 50,000. The results in Table \ref{tab:changingannotationsize} show that HUDS consistently shows a higher \textsc{SacreBLEU} score than other methods with a larger annotation size.

\begin{table}[H]
\centering

\resizebox{0.90\columnwidth}{!}{
\begin{tabular}{c c c c c c c}
%\small
\hline
Method & \multicolumn{5}{c}{Iteration}\\
\hline
\hline
{} & \textbf{2} & \textbf{4} & \textbf{6}  & \textbf{8}  & \textbf{10} \\
\hline

Random   & 29.7   & 30.7   & 31.5   & 31.9   & 32.5    \\
IDDS     & 25.6   & 27.1   & 27.9   & 28.7   & 29.1    \\ 
NSP      & 30.5   & 31.2   & 31.9   & 32.0   & 32.3    \\ 
HUDS     & \textbf{30.8}   & \textbf{32.3}   & \textbf{32.9}   & \textbf{33.4}   & \textbf{34.0}    \\ \hline

\hline
\end{tabular}}
\caption{Mean \textsc{SacreBLEU} score on the IT dataset. 5000 examples are selected for labeling in each iteration. }
\label{tab:changingannotationsize}
% \vskip -0.2in
\end{table}

\subsection{Composition of annotated sentences}

To find out how the composition of annotated instances affects the generalization performance of the NMT model, it is useful to analyze the n-gram overlap between the selected sentences and the test set. Prior work \cite{hu-neubig-2021-phrase} has shown that the n-gram overlap correlates highly with test BLEU scores, making it a reliable measure of the usefulness of selected data for domain adaptation. Thus, we conduct an experiment to analyze the unigram overlap between the instances selected by different methods and the validation set of IT corpus, following the protocol presented in \citet{hu-neubig-2021-phrase}. We find that the sentences selected by HUDS have the highest overall unigram overlap with the validation set, followed by random, NSP, and IDDS (Table \ref{tab:overlap}). This indicates that HUDS selects varied instances with higher unigram coverage, potentially leading to better performance. We also compare the overlap of English and German individually, which shows a similar trend, except that NSP has a higher English unigram overlap than the randomly annotated sentences (46.7 vs 46.6).

\begin{table}[!htbp]
\centering
\resizebox{1\columnwidth}{!}{%
\begin{tabular}{lllll}
\hline Method & \ En Overlap (\%) &  De Overlap (\%) & En-De Overlap (\%) \\
\hline 
HUDS  & \textbf{47.2} & \textbf{53.2} & \textbf{48.0}  \\
IDDS   & 46.0 & 52.1 &  47.3 \\
NSP    & 46.7 & 52.2 & 47.6   \\
Random   & 46.6 & 53.0 & 47.8 \\
\hline
\end{tabular}}
\caption{Percentage overlap between the unigrams of sentences selected by different AL strategies and the validation set of IT corpus. We report the individual overlap for En and De and the En-De combined overlap.} 
\label{tab:overlap}
\vskip -0.2in

\end{table}

\subsection{COMET scores}
We additionally assess the translation quality of HUDS using COMET \cite{rei-etal-2020-comet}, a state-of-the-art neural evaluation metric that demonstrates high correlation with human judgment. We use the \texttt{Unbabel/wmt20-comet-da} model\footnote{\unbabelcomet} for scoring the translations on the IT domain dataset. The results in Table \ref{tab:cometscores} show that HUDS outperforms other active learning strategies.

\begin{table}[H]
\centering

\resizebox{0.7\columnwidth}{!}{
\begin{tabular}{c c c c}
%\small
\hline
\textbf{Random} & \textbf{IDDS} & \textbf{NSP} & \textbf{HUDS} \\
\hline
42.5  & 38.0 & 42.3 & \textbf{42.6} \\
\hline
\end{tabular}}

\caption{Comparison of the \textsc{COMET} scores after the final AL iteration for different AL strategies on the medical domain dataset. \texttt{Unbabel/wmt20-comet-da} model is used for evaluation. Active learning is done over ten iterations, with 1000 sentences selected in each iteration.}

\label{tab:cometscores}
\end{table}

\section{Ablation study}
% \subsection{The impact of parameter $\lambda$ in HUDS}
An important hyperparameter in the HUDS algorithm is the parameter $\lambda$ that balances uncertainty and diversity score for a sentence. We now study the effect of varying the value of $\lambda$ in HUDS on the \textsc{SacreBLEU} score for the multi-domains dataset. We run a separate AL procedure for each value of $\lambda$ on different datasets and report  \textsc{SacreBLEU} score at the end of the first iteration with 1000 sentences annotated and used for training in that iteration. The ablation is \textit{done on a single iteration} to better isolate the impact of individual components and avoid any potential confounding effects. The results in Table \ref{tab:changinglambda} demonstrate a higher \textsc{SacreBLEU} score for $\lambda = 0.5$, indicating that a balance between hybrid uncertainty and diversity scores in HUDS leads to better generalization. Additionally, the parameter $\lambda$ provides a powerful tuning method for uncertainty or diversity sampling using just the validation set.

\begin{table}[!htbp]
\centering

\resizebox{\columnwidth}{!}{
\begin{tabular}{c c c c c c c c}
%\small
\hline
Dataset & \multicolumn{7}{c}{$\lambda$}\\
\hline
\hline
{} & \textbf{0}  & \textbf{0.1} & \textbf{0.3} & \textbf{0.5}  & \textbf{0.7}  & \textbf{0.9}  & \textbf{1.0} \\
\hline
Medicine  & 29.6 &  29.6 & 29.7 & \textbf{29.9} & 29.7 & 29.7  & 29.5  \\
Law  & 32.8  & 32.8  &  32.7 & \textbf{33.0} & 	32.8 &	32.9  & 31.2	\\
IT & 28.7 & 28.8  &	28.6  &	\textbf{29.0}  &	28.7 &	28.6  & 28.5\\

\hline
\end{tabular}}
\caption{\textsc{SacreBLEU} score after the first AL iteration for different values of $\lambda$ in HUDS.}
\label{tab:changinglambda}
\vskip -0.2in
\end{table}

\section{Related Work}

\textbf{Active Learning in Machine Translation.} Active Learning has been widely applied to machine translation (MT) to improve the quality and efficiency of translation systems, especially for low-resource languages or domains  \cite{ zhang2018active, hu-neubig-2021-phrase, gupta2021investigating,  zhou2021active, zhao2020active}. Several aspects of AL have been explored for MT, including different sampling techniques that consider uncertainty or diversity \cite{ambati2011multi}, different workflows including batch mode vs. online mode \cite{ananthakrishnan-etal-2010-semi, lam-etal-2019-interactive}, leveraging pre-trained or auxiliary models \cite{domhan-hieber-2017-using,wang-etal-2020-balancing} and exploiting domain-specific and synthetic data \cite{zhang-etal-2019-curriculum,hoang-etal-2018-iterative}. To our knowledge, no hybrid AL strategy for efficiently acquiring domain-specific data in NMT has been proposed prior to our work.

\medskip

\noindent \textbf{Uncertainty Sampling Techniques.} Some of the early works on AL for MT use uncertainty-based sampling criteria, such as entropy or posterior probability, to select the sentences that the model is most uncertain about for annotation \cite{haffari2009active,bloodgood-vijay-shanker-2009-taking, zhao2020active}. Using these sentences will enable the model to generalize better to unseen data. For NMT models, uncertainty is usually measured at the sentence or the beam level, based on the output of an encoder-decoder model or a transformer model \cite{zhou-etal-2020-uncertainty}. However, pure uncertainty sampling techniques suffer from redundancy issues as they tend to select uncertain similar sentences or prefer certain words and structures.

\medskip
\noindent \textbf{Diversity Sampling Techniques.} The main idea of diversity sampling is to select the sentences that are most diverse and representative of the unlabeled data based on some diversity measures such as density, representativeness, or coverage \cite{bloodgood-callison-burch-2010-bucking, gangadharaiah-etal-2009-active, hu-neubig-2021-phrase, pendas2023neural}. Diversity is usually measured at the word or phrase level based on the frequency or similarity of words or phrases. However, pure diversity sampling methods lead to the selection of noisier or irrelevant instances that are too difficult or too easy for the model.

\section{Conclusion}

In this work, we proposed HUDS, a novel hybrid active learning strategy that leverages both uncertainty and diversity for neural machine translation. HUDS outperforms other AL methods for NMT, demonstrating the effectiveness of hybrid sampling over other baselines for the same amount of human labeling effort. Future work includes (1) incorporating phrasal selection in HUDS and (2) making \textsc{HUDS} more efficient through precomputation of embeddings and intermediate caching of scores.  

\section{Limitations}
We list the potential limitations of our work below: (1) HUDS involves computation of embeddings, clustering, and calculation of hybrid sampling scores for each unlabeled sentence, which could lead to higher latency compared to other AL methods. For large datasets, the additional computations may affect the \textit{interactivity} of the AL procedure. Precomputation of embeddings and intermediate caching of scores can be leveraged to mitigate this. HUDS can also be augmented with automatic quality estimation methods for NMT \cite{blain-etal-2023-findings} to increase efficiency. (2) The proposed hybrid sampling strategy HUDS focuses on sentence-level active learning for NMT. Phrase-level active learning could potentially be more cost-effective by selecting only informative phrases within sentences. Extending HUDS to phrase-level selection is left for future work.

\bibliography{anthology,custom}

\begin{thebibliography}{43}
\providecommand{\natexlab}[1]{#1}

\bibitem[{Aharoni and Goldberg(2020)}]{aharoni-goldberg-2020-unsupervised}
Roee Aharoni and Yoav Goldberg. 2020.
\newblock \href {https://doi.org/10.18653/v1/2020.acl-main.692} {Unsupervised domain clusters in pretrained language models}.
\newblock In \emph{Proceedings of the 58th Annual Meeting of the Association for Computational Linguistics}, pages 7747--7763, Online. Association for Computational Linguistics.

\bibitem[{Ambati et~al.(2011)Ambati, Vogel, and Carbonell}]{ambati2011multi}
Vamshi Ambati, Stephan Vogel, and Jaime~G Carbonell. 2011.
\newblock Multi-strategy approaches to active learning for statistical machine translation.
\newblock In \emph{Proceedings of Machine Translation Summit XIII: Papers}.

\bibitem[{Ananthakrishnan et~al.(2010)Ananthakrishnan, Prasad, Stallard, and Natarajan}]{ananthakrishnan-etal-2010-semi}
Sankaranarayanan Ananthakrishnan, Rohit Prasad, David Stallard, and Prem Natarajan. 2010.
\newblock \href {https://aclanthology.org/W10-2916} {A semi-supervised batch-mode active learning strategy for improved statistical machine translation}.
\newblock In \emph{Proceedings of the Fourteenth Conference on Computational Natural Language Learning}, pages 126--134, Uppsala, Sweden. Association for Computational Linguistics.

\bibitem[{Arthur and Vassilvitskii(2007)}]{arthur2007k}
David Arthur and Sergei Vassilvitskii. 2007.
\newblock K-means++ the advantages of careful seeding.
\newblock In \emph{Proceedings of the eighteenth annual ACM-SIAM symposium on Discrete algorithms}, pages 1027--1035.

\bibitem[{Ash et~al.(2019)Ash, Zhang, Krishnamurthy, Langford, and Agarwal}]{ash2019deep}
Jordan~T Ash, Chicheng Zhang, Akshay Krishnamurthy, John Langford, and Alekh Agarwal. 2019.
\newblock Deep batch active learning by diverse, uncertain gradient lower bounds.
\newblock \emph{arXiv preprint arXiv:1906.03671}.

\bibitem[{Blain et~al.(2023)Blain, Zerva, Rei, Guerreiro, Kanojia, C.~de Souza, Silva, Vaz, Jingxuan, Azadi, Orasan, and Martins}]{blain-etal-2023-findings}
Frederic Blain, Chrysoula Zerva, Ricardo Rei, Nuno~M. Guerreiro, Diptesh Kanojia, Jos{\'e}~G. C.~de Souza, Beatriz Silva, T{\^a}nia Vaz, Yan Jingxuan, Fatemeh Azadi, Constantin Orasan, and Andr{\'e} Martins. 2023.
\newblock \href {https://doi.org/10.18653/v1/2023.wmt-1.52} {Findings of the {WMT} 2023 shared task on quality estimation}.
\newblock In \emph{Proceedings of the Eighth Conference on Machine Translation}, pages 629--653, Singapore. Association for Computational Linguistics.

\bibitem[{Bloodgood and Callison-Burch(2010)}]{bloodgood-callison-burch-2010-bucking}
Michael Bloodgood and Chris Callison-Burch. 2010.
\newblock \href {https://aclanthology.org/P10-1088} {Bucking the trend: Large-scale cost-focused active learning for statistical machine translation}.
\newblock In \emph{Proceedings of the 48th Annual Meeting of the Association for Computational Linguistics}, pages 854--864, Uppsala, Sweden. Association for Computational Linguistics.

\bibitem[{Bloodgood and Vijay-Shanker(2009)}]{bloodgood-vijay-shanker-2009-taking}
Michael Bloodgood and K.~Vijay-Shanker. 2009.
\newblock \href {https://aclanthology.org/N09-2035} {Taking into account the differences between actively and passively acquired data: The case of active learning with support vector machines for imbalanced datasets}.
\newblock In \emph{Proceedings of Human Language Technologies: The 2009 Annual Conference of the North {A}merican Chapter of the Association for Computational Linguistics, Companion Volume: Short Papers}, pages 137--140, Boulder, Colorado. Association for Computational Linguistics.

\bibitem[{Dasgupta(2011)}]{dasgupta2011two}
Sanjoy Dasgupta. 2011.
\newblock Two faces of active learning.
\newblock \emph{Theoretical computer science}, 412(19):1767--1781.

\bibitem[{Domhan and Hieber(2017)}]{domhan-hieber-2017-using}
Tobias Domhan and Felix Hieber. 2017.
\newblock \href {https://doi.org/10.18653/v1/D17-1158} {Using target-side monolingual data for neural machine translation through multi-task learning}.
\newblock In \emph{Proceedings of the 2017 Conference on Empirical Methods in Natural Language Processing}, pages 1500--1505, Copenhagen, Denmark. Association for Computational Linguistics.

\bibitem[{Dor et~al.(2020)Dor, Halfon, Gera, Shnarch, Dankin, Choshen, Danilevsky, Aharonov, Katz, and Slonim}]{dor2020active}
Liat~Ein Dor, Alon Halfon, Ariel Gera, Eyal Shnarch, Lena Dankin, Leshem Choshen, Marina Danilevsky, Ranit Aharonov, Yoav Katz, and Noam Slonim. 2020.
\newblock Active learning for bert: an empirical study.
\newblock In \emph{Proceedings of the 2020 Conference on Empirical Methods in Natural Language Processing (EMNLP)}, pages 7949--7962.

\bibitem[{Dou et~al.(2019)Dou, Hu, Anastasopoulos, and Neubig}]{dou19emnlp}
Zi-Yi Dou, Junjie Hu, Antonios Anastasopoulos, and Graham Neubig. 2019.
\newblock Unsupervised domain adaptation for neural machine translation with domain-aware feature embeddings.
\newblock In \emph{Conference on Empirical Methods in Natural Language Processing (EMNLP)}, Hong Kong.

\bibitem[{Gangadharaiah et~al.(2009)Gangadharaiah, Brown, and Carbonell}]{gangadharaiah-etal-2009-active}
Rashmi Gangadharaiah, Ralf~D. Brown, and Jaime Carbonell. 2009.
\newblock \href {https://aclanthology.org/W09-4633} {Active learning in example-based machine translation}.
\newblock In \emph{Proceedings of the 17th Nordic Conference of Computational Linguistics ({NODALIDA} 2009)}, pages 227--230, Odense, Denmark. Northern European Association for Language Technology (NEALT).

\bibitem[{Gupta et~al.(2021)Gupta, Boppana, Haque, Ekbal, and Bhattacharyya}]{gupta2021investigating}
Kamal Gupta, Dhanvanth Boppana, Rejwanul Haque, Asif Ekbal, and Pushpak Bhattacharyya. 2021.
\newblock Investigating active learning in interactive neural machine translation.
\newblock In \emph{Proceedings of Machine Translation Summit XVIII: Research Track}, pages 10--22.

\bibitem[{Haffari et~al.(2009)Haffari, Roy, and Sarkar}]{haffari2009active}
Gholamreza Haffari, Maxim Roy, and Anoop Sarkar. 2009.
\newblock \href {https://aclanthology.org/N09-1047} {Active learning for statistical phrase-based machine translation}.
\newblock In \emph{Proceedings of Human Language Technologies: The 2009 Annual Conference of the North {A}merican Chapter of the Association for Computational Linguistics}, pages 415--423, Boulder, Colorado. Association for Computational Linguistics.

\bibitem[{Hoang et~al.(2018)Hoang, Koehn, Haffari, and Cohn}]{hoang-etal-2018-iterative}
Vu~Cong~Duy Hoang, Philipp Koehn, Gholamreza Haffari, and Trevor Cohn. 2018.
\newblock \href {https://doi.org/10.18653/v1/W18-2703} {Iterative back-translation for neural machine translation}.
\newblock In \emph{Proceedings of the 2nd Workshop on Neural Machine Translation and Generation}, pages 18--24, Melbourne, Australia. Association for Computational Linguistics.

\bibitem[{Hsu and Lin(2015)}]{hsu2015active}
Wei-Ning Hsu and Hsuan-Tien Lin. 2015.
\newblock Active learning by learning.
\newblock In \emph{Proceedings of the AAAI Conference on Artificial Intelligence}, volume~29.

\bibitem[{Hu and Neubig(2021{\natexlab{a}})}]{hu-neubig-2021-phrase}
Junjie Hu and Graham Neubig. 2021{\natexlab{a}}.
\newblock \href {https://aclanthology.org/2021.wmt-1.117} {Phrase-level active learning for neural machine translation}.
\newblock In \emph{Proceedings of the Sixth Conference on Machine Translation}, pages 1087--1099, Online. Association for Computational Linguistics.

\bibitem[{Hu and Neubig(2021{\natexlab{b}})}]{hu2021phrase}
Junjie Hu and Graham Neubig. 2021{\natexlab{b}}.
\newblock \href {https://aclanthology.org/2021.wmt-1.117} {Phrase-level active learning for neural machine translation}.
\newblock pages 1087--1099.

\bibitem[{Kim et~al.(2006)Kim, Song, Kim, Cha, and Lee}]{kim-etal-2006-mmr}
Seokhwan Kim, Yu~Song, Kyungduk Kim, Jeong-Won Cha, and Gary~Geunbae Lee. 2006.
\newblock \href {https://aclanthology.org/N06-2018} {{MMR}-based active machine learning for bio named entity recognition}.
\newblock In \emph{Proceedings of the Human Language Technology Conference of the {NAACL}, Companion Volume: Short Papers}, pages 69--72, New York City, USA. Association for Computational Linguistics.

\bibitem[{Koehn and Knowles(2017)}]{koehn-knowles-2017-six}
Philipp Koehn and Rebecca Knowles. 2017.
\newblock \href {https://doi.org/10.18653/v1/W17-3204} {Six challenges for neural machine translation}.
\newblock In \emph{Proceedings of the First Workshop on Neural Machine Translation}, pages 28--39, Vancouver. Association for Computational Linguistics.

\bibitem[{Kothawade et~al.(2022)Kothawade, Kaushal, Ramakrishnan, Bilmes, and Iyer}]{kothawade2022prism}
Suraj Kothawade, Vishal Kaushal, Ganesh Ramakrishnan, Jeff Bilmes, and Rishabh Iyer. 2022.
\newblock Prism: A rich class of parameterized submodular information measures for guided data subset selection.
\newblock In \emph{Proceedings of the AAAI Conference on Artificial Intelligence}, volume~36, pages 10238--10246.

\bibitem[{Lam et~al.(2019)Lam, Schamoni, and Riezler}]{lam-etal-2019-interactive}
Tsz~Kin Lam, Shigehiko Schamoni, and Stefan Riezler. 2019.
\newblock \href {https://aclanthology.org/W19-6610} {Interactive-predictive neural machine translation through reinforcement and imitation}.
\newblock In \emph{Proceedings of Machine Translation Summit XVII: Research Track}, pages 96--106, Dublin, Ireland. European Association for Machine Translation.

\bibitem[{Lewis et~al.(2020)Lewis, Liu, Goyal, Ghazvininejad, Mohamed, Levy, Stoyanov, and Zettlemoyer}]{lewis-etal-2020-bart}
Mike Lewis, Yinhan Liu, Naman Goyal, Marjan Ghazvininejad, Abdelrahman Mohamed, Omer Levy, Veselin Stoyanov, and Luke Zettlemoyer. 2020.
\newblock \href {https://doi.org/10.18653/v1/2020.acl-main.703} {{BART}: Denoising sequence-to-sequence pre-training for natural language generation, translation, and comprehension}.
\newblock In \emph{Proceedings of the 58th Annual Meeting of the Association for Computational Linguistics}, pages 7871--7880, Online. Association for Computational Linguistics.

\bibitem[{Miura et~al.(2016)Miura, Neubig, Paul, and Nakamura}]{miura2016selecting}
Akiva Miura, Graham Neubig, Michael Paul, and Satoshi Nakamura. 2016.
\newblock Selecting syntactic, non-redundant segments in active learning for machine translation.
\newblock In \emph{Proceedings of the 2016 Conference of the North American Chapter of the Association for Computational Linguistics: Human Language Technologies}, pages 20--29.

\bibitem[{Pendas et~al.(2023)Pendas, Carvallo, and Aspillaga}]{pendas2023neural}
Begoa Pendas, Andr{\'e}s Carvallo, and Carlos Aspillaga. 2023.
\newblock Neural machine translation through active learning on low-resource languages: The case of spanish to mapudungun.
\newblock In \emph{Proceedings of the Workshop on Natural Language Processing for Indigenous Languages of the Americas (AmericasNLP)}, pages 6--11.

\bibitem[{Post(2018)}]{post-2018-call}
Matt Post. 2018.
\newblock \href {https://doi.org/10.18653/v1/W18-6319} {A call for clarity in reporting {BLEU} scores}.
\newblock In \emph{Proceedings of the Third Conference on Machine Translation: Research Papers}, pages 186--191, Brussels, Belgium. Association for Computational Linguistics.

\bibitem[{Rei et~al.(2020)Rei, Stewart, Farinha, and Lavie}]{rei-etal-2020-comet}
Ricardo Rei, Craig Stewart, Ana~C Farinha, and Alon Lavie. 2020.
\newblock \href {https://doi.org/10.18653/v1/2020.emnlp-main.213} {{COMET}: A neural framework for {MT} evaluation}.
\newblock In \emph{Proceedings of the 2020 Conference on Empirical Methods in Natural Language Processing (EMNLP)}, pages 2685--2702, Online. Association for Computational Linguistics.

\bibitem[{Sainz et~al.(2023)Sainz, Campos, Garc{\'\i}a-Ferrero, Etxaniz, de~Lacalle, and Agirre}]{sainz-etal-2023-nlp}
Oscar Sainz, Jon Campos, Iker Garc{\'\i}a-Ferrero, Julen Etxaniz, Oier~Lopez de~Lacalle, and Eneko Agirre. 2023.
\newblock \href {https://doi.org/10.18653/v1/2023.findings-emnlp.722} {{NLP} evaluation in trouble: On the need to measure {LLM} data contamination for each benchmark}.
\newblock In \emph{Findings of the Association for Computational Linguistics: EMNLP 2023}, pages 10776--10787, Singapore. Association for Computational Linguistics.

\bibitem[{Shelmanov et~al.(2021)Shelmanov, Puzyrev, Kupriyanova, Belyakov, Larionov, Khromov, Kozlova, Artemova, Dylov, and Panchenko}]{shelmanov_active}
Artem Shelmanov, Dmitri Puzyrev, Lyubov Kupriyanova, Denis Belyakov, Daniil Larionov, Nikita Khromov, Olga Kozlova, Ekaterina Artemova, Dmitry~V. Dylov, and Alexander Panchenko. 2021.
\newblock \href {https://www.aclweb.org/anthology/2021.eacl-main.145} {Active learning for sequence tagging with deep pre-trained models and {B}ayesian uncertainty estimates}.
\newblock In \emph{Proceedings of the 16th Conference of the European Chapter of the Association for Computational Linguistics: Main Volume}, pages 1698--1712, Online. Association for Computational Linguistics.

\bibitem[{Tsvigun et~al.(2022{\natexlab{a}})Tsvigun, Lysenko, Sedashov, Lazichny, Damirov, Karlov, Belousov, Sanochkin, Panov, Panchenko, Burtsev, and Shelmanov}]{tsvigun2023active}
Akim Tsvigun, Ivan Lysenko, Danila Sedashov, Ivan Lazichny, Eldar Damirov, Vladimir Karlov, Artemy Belousov, Leonid Sanochkin, Maxim Panov, Alexander Panchenko, Mikhail Burtsev, and Artem Shelmanov. 2022{\natexlab{a}}.
\newblock \href {https://doi.org/10.18653/v1/2022.findings-emnlp.377} {Active learning for abstractive text summarization}.
\newblock In \emph{Findings of the Association for Computational Linguistics: EMNLP 2022}, pages 5128--5152, Abu Dhabi, United Arab Emirates. Association for Computational Linguistics.

\bibitem[{Tsvigun et~al.(2022{\natexlab{b}})Tsvigun, Sanochkin, Larionov, Kuzmin, Vazhentsev, Lazichny, Khromov, Kireev, Rubashevskii, Shahmatova, Dylov, Galitskiy, and Shelmanov}]{tsvigun-etal-2022-altoolbox}
Akim Tsvigun, Leonid Sanochkin, Daniil Larionov, Gleb Kuzmin, Artem Vazhentsev, Ivan Lazichny, Nikita Khromov, Danil Kireev, Aleksandr Rubashevskii, Olga Shahmatova, Dmitry~V. Dylov, Igor Galitskiy, and Artem Shelmanov. 2022{\natexlab{b}}.
\newblock \href {https://doi.org/10.18653/v1/2022.emnlp-demos.41} {{ALT}oolbox: A set of tools for active learning annotation of natural language texts}.
\newblock In \emph{Proceedings of the 2022 Conference on Empirical Methods in Natural Language Processing: System Demonstrations}, pages 406--434, Abu Dhabi, UAE. Association for Computational Linguistics.

\bibitem[{Ueffing and Ney(2007)}]{ueffing_sp}
Nicola Ueffing and Hermann Ney. 2007.
\newblock \href {https://doi.org/10.1162/coli.2007.33.1.9} {Word-level confidence estimation for machine translation}.
\newblock \emph{Comput. Linguistics}, 33(1):9--40.

\bibitem[{Wang et~al.(2017)Wang, Finch, Utiyama, and Sumita}]{wang2017sentence}
Rui Wang, Andrew Finch, Masao Utiyama, and Eiichiro Sumita. 2017.
\newblock Sentence embedding for neural machine translation domain adaptation.
\newblock In \emph{Proceedings of the 55th Annual Meeting of the Association for Computational Linguistics (Volume 2: Short Papers)}, pages 560--566.

\bibitem[{Wang et~al.(2020)Wang, Tsvetkov, and Neubig}]{wang-etal-2020-balancing}
Xinyi Wang, Yulia Tsvetkov, and Graham Neubig. 2020.
\newblock \href {https://doi.org/10.18653/v1/2020.acl-main.754} {Balancing training for multilingual neural machine translation}.
\newblock In \emph{Proceedings of the 58th Annual Meeting of the Association for Computational Linguistics}, pages 8526--8537, Online. Association for Computational Linguistics.

\bibitem[{Wolf et~al.(2019)Wolf, Debut, Sanh, Chaumond, Delangue, Moi, Cistac, Rault, Louf, Funtowicz et~al.}]{wolf2019huggingface}
Thomas Wolf, Lysandre Debut, Victor Sanh, Julien Chaumond, Clement Delangue, Anthony Moi, Pierric Cistac, Tim Rault, R{\'e}mi Louf, Morgan Funtowicz, et~al. 2019.
\newblock Huggingface's transformers: State-of-the-art natural language processing.
\newblock \emph{arXiv preprint arXiv:1910.03771}.

\bibitem[{Xia et~al.(2024)Xia, Liu, Yu, Kim, Rossi, Rao, Mai, and Li}]{xia-etal-2024-hallucination}
Yu~Xia, Xu~Liu, Tong Yu, Sungchul Kim, Ryan Rossi, Anup Rao, Tung Mai, and Shuai Li. 2024.
\newblock \href {https://doi.org/10.18653/v1/2024.naacl-long.479} {Hallucination diversity-aware active learning for text summarization}.
\newblock In \emph{Proceedings of the 2024 Conference of the North American Chapter of the Association for Computational Linguistics: Human Language Technologies (Volume 1: Long Papers)}, pages 8665--8677, Mexico City, Mexico. Association for Computational Linguistics.

\bibitem[{Yuan et~al.(2020)Yuan, Lin, and Boyd-Graber}]{yuan-etal-2020-cold}
Michelle Yuan, Hsuan-Tien Lin, and Jordan Boyd-Graber. 2020.
\newblock \href {https://doi.org/10.18653/v1/2020.emnlp-main.637} {Cold-start active learning through self-supervised language modeling}.
\newblock In \emph{Proceedings of the 2020 Conference on Empirical Methods in Natural Language Processing (EMNLP)}, pages 7935--7948, Online. Association for Computational Linguistics.

\bibitem[{Zhang et~al.(2018)Zhang, Xu, and Xiong}]{zhang2018active}
Pei Zhang, Xueying Xu, and Deyi Xiong. 2018.
\newblock Active learning for neural machine translation.
\newblock In \emph{2018 International Conference on Asian Language Processing (IALP)}, pages 153--158. IEEE.

\bibitem[{Zhang et~al.(2019)Zhang, Shapiro, Kumar, McNamee, Carpuat, and Duh}]{zhang-etal-2019-curriculum}
Xuan Zhang, Pamela Shapiro, Gaurav Kumar, Paul McNamee, Marine Carpuat, and Kevin Duh. 2019.
\newblock \href {https://doi.org/10.18653/v1/N19-1189} {Curriculum learning for domain adaptation in neural machine translation}.
\newblock In \emph{Proceedings of the 2019 Conference of the North {A}merican Chapter of the Association for Computational Linguistics: Human Language Technologies, Volume 1 (Long and Short Papers)}, pages 1903--1915, Minneapolis, Minnesota. Association for Computational Linguistics.

\bibitem[{Zhao et~al.(2020)Zhao, Zhang, Zhou, and Zhang}]{zhao2020active}
Yuekai Zhao, Haoran Zhang, Shuchang Zhou, and Zhihua Zhang. 2020.
\newblock Active learning approaches to enhancing neural machine translation.
\newblock In \emph{Findings of the Association for Computational Linguistics: EMNLP 2020}, pages 1796--1806.

\bibitem[{Zhou et~al.(2020)Zhou, Yang, Wong, Wan, and Chao}]{zhou-etal-2020-uncertainty}
Yikai Zhou, Baosong Yang, Derek~F. Wong, Yu~Wan, and Lidia~S. Chao. 2020.
\newblock \href {https://doi.org/10.18653/v1/2020.acl-main.620} {Uncertainty-aware curriculum learning for neural machine translation}.
\newblock In \emph{Proceedings of the 58th Annual Meeting of the Association for Computational Linguistics}, pages 6934--6944, Online. Association for Computational Linguistics.

\bibitem[{Zhou and Waibel(2021)}]{zhou2021active}
Zhong Zhou and Alex Waibel. 2021.
\newblock Active learning for massively parallel translation of constrained text into low resource languages.
\newblock \emph{arXiv preprint arXiv:2108.07127}.

\end{thebibliography}

\appendix

\section{Implementation Details}
We extend the \texttt{ALToolbox}\footnote{\altoolbox} python package \cite{tsvigun-etal-2022-altoolbox} and add support for hybrid uncertainty and diversity sampling (HUDS) for active learning (AL). The implementation of HUDS allows using any translation dataset\footnote{\huggingfacedatasets} and model\footnote{\huggingfacemodels} available on HuggingFace \cite{wolf2019huggingface}. The code is submitted as part of the supplementary material.

\subsection{Models}
We use the \texttt{BART-base} model publicly available on HuggingFace transformers \citep{wolf2019huggingface}. The HuggingFace repository is available under the Apache License 2.0 license.

\subsection{Datasets}
We use the WMT14 \texttt{En-De} dataset available under the CC-BY-SA license and the German-English multi-domains parallel data available under the Creative Commons CC0 license.

\section{Hyperparameters}
The hyperparameters used in our experiments are reported in Table \ref{tab:hyperparameters}. We use the validation set of the multi-domains dataset (Medicine, IT, and Law) to tune these parameters.

\begin{table}[H]
\centering
\resizebox{0.8\columnwidth}{!}{
\begin{tabular}{lc}
\hline Hyperparameter & Value \\
\hline
learning rate & $2 e-5$ \\
train batch size & $16$ \\
eval batch size & $16$ \\
num beams & $4$ \\
pad-to-max-length & true \\
gradient-clipping & $0.28$ \\
scheduler warmup-steps-factor & 0.1 \\
weight-decay & 0.028 \\
gradient-accumulation & 0.1 \\
fp-16 & true \\
number of strata & 10 \\

\hline
\end{tabular}
}

\caption{Hyperparameters for the experiments. } 
\label{tab:hyperparameters}
\end{table}

\section{Resources}
The experiments are run on a single 48GB NVIDIA A6000 GPU on the cloud. Around 1680 GPU hours were used for the entirety of the project which includes the early experiments and the final results.

\end{document}